\documentclass{article}

\usepackage{arxiv}

\usepackage[utf8]{inputenc} 
\usepackage[T1]{fontenc}    
\usepackage{hyperref}       
\usepackage{url}            
\usepackage{booktabs}       
\usepackage{amsfonts}       
\usepackage{nicefrac}       
\usepackage{microtype}      
\usepackage{lipsum}		
\usepackage{graphicx}
\usepackage{cite}
\usepackage{doi}
\usepackage{amssymb}
\usepackage{tabularx}
\usepackage{ltablex}

\title{Credit Card Fraud Detection using Machine Learning: A Study}

\author{Pooja Tiwari \\
	Department of Computer Applications\\
	National Institute of Technology\\
	Kurukshetra, Haryana 136119 \\
	\texttt{pooja\_51710011@nitkkr.ac.in} \\
	\AND 
	Simran Mehta\\
	Department of Computer Applications\\
	National Institute of Technology\\
	Kurukshetra, Haryana 136119 \\
	\texttt{simran\_51710060@nitkkr.ac.in} \\
	\AND 
	Nishtha Sakhuja\\
	Department of Computer Applications\\
	National Institute of Technology\\
	Kurukshetra, Haryana 136119 \\
	\texttt{nishtha\_51710077@nitkkr.ac.in} \\
	\AND
	\href{https://orcid.org/0000-0002-2938-6432}{\includegraphics[scale=0.06]{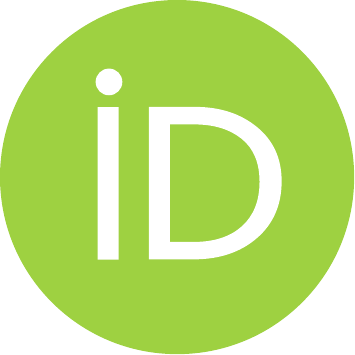}\hspace{1mm}Jitendra Kumar} \\
	Department of Computer Applications\\
	National Institute of Technology\\
	Tiruchirappalli, Tamilnadu 620015 \\
	\texttt{jitendra@nitt.edu} \\
	\AND 
	\href{https://orcid.org/0000-0002-8053-5050}{\includegraphics[scale=0.06]{orcid.pdf}\hspace{1mm}Ashutosh Kumar Singh}\\
	Department of Computer Applications\\
	National Institute of Technology\\
	Kurukshetra, Haryana 136119 \\
	\texttt{ashutosh@nitkkr.ac.in} \\
}

\date{}

\renewcommand{\headeright}{Technical Report}
\renewcommand{\undertitle}{Technical Report}
\renewcommand{\shorttitle}{Credit Card Fraud Detection using Machine Learning: A Study}

\hypersetup{
pdftitle={Credit Card Fraud Detection using Machine Learning: A Study},
pdfauthor={Jitendra Kumar},
pdfkeywords={Hidden Markov Model, Decision Trees, Logistic Regression, Support Vector Machines (SVM), Genetic algorithm, Neural Networks, Random Forests, Bayesian Belief Network},
}

\begin{document}
\maketitle

\begin{abstract}
	As the world is rapidly moving towards digitization and money transactions are becoming cashless, the use of credit cards has rapidly increased. The fraud activities associated with it have also been increasing which leads to a huge loss to the financial institutions. Therefore, we need to analyze and detect the fraudulent transaction from the non-fraudulent ones. In this paper, we present a comprehensive review of various methods used to detect credit card frauds. These methodologies include Hidden Markov Model, Decision Trees, Logistic Regression, Support Vector Machines (SVM), Genetic algorithm, Neural Networks, Random Forests, Bayesian Belief Network. A comprehensive analysis of various techniques is presented. We conclude the paper with the pros and cons of the same as stated in the respective papers.
\end{abstract}


\section{Introduction}
Credit card being one of the most used financial products is designed to make purchases such as gas, groceries, TVs, traveling, shopping bills and so on because of non-availability of funds at that instance. Credit cards are of most value that provides various benefits in the form of points while using them for different types of transactions. Usually major hotels as well as various car-renting firms require the buyer to own a credit card for the same. Towards the end of 2005, huge amounts of sales were generated that is approximately around \$190.6 billion just by circulating around 56.4 million credit cards in Canada. Means of achieving money or services and goods by illegal or unethical means is said to be a fraud. Banking fraud is basically ``The unauthorized use of an individual’s confidential information to make purchases, or to remove funds from the user’s account.'' According to survey of statista\cite{2}, there were about 792.6 million of digital consumers in around 2011 with the number rising to 903.6 million in the following year. In 2013, around 41.3\% of users that use internet started buying products online, this expecting to having reached 46.4\% by 2017. With e-commerce increasing rapidly, and with the world moving towards digitization, towards cashless transactions, use of credit-card users has increased rapidly, and with that, the number of frauds associated with it is also increasing. 

There are several categories of credit card frauds that are observed:

\subsection{Card/Account Holder}
The ``base level'' of fraud activity accounts for the lost and stolen card and the economic conditions size of this base-level (e.g., high levels of unemployment lead to increase in fraud because of lost and stolen cards). Counterfeit cards fraud, being a more structured problem in comparison with stolen and lost cards, is a growing problem, even after sophisticated card manufacturing technologies available like the presence of holograms and magnetic stripes on the card with some information to encrypt.
 
Another category of fraud of cards that is stealing cards from the mail, the non-receipt of issue (NRI) fraud affects buyers issuing both new and the ones for re-issues and this problem has increased so severely that some issuers have started using other card delivery options like courier instead of mail and also special card activation programs in which bank keeps the card blocked (the ones which are included in list as an account whose transaction requests will be denied) until the customer verifies card receipt. The customer calls and bank use it to account for the fact that he caller is genuine cardholder by asking few questions about the background of the customer using the card application or its cardholder information file and these have led to a decrease in NRI losses.

Other frauds include sending of illegitimate applications for a card because of which criminals send applications using one’s personal background and financial information to specify a mail for receiving the card. In such cases, even the activation of the card member cannot stop the card being misused and fall into wrong hands. Another type of fraud is mail-orded telephone order fraud in which card imprint cannot be obtained as the buyer is not present there during the transaction.

\subsection{Merchant Fraud}
Another type of fraud is the merchant fraud involving the ``laundering'' of merchant receipts gaining huge amounts by showing illegal transactions that never occurred \cite{22}. Many a times, during admissible transactions, legal information is stolen which is then used to illegally produce copies of cards to perform illegal transactions. Here the dealer is a ``point of compromise'' as all these cards have transaction histories that establishing the dealer (merchant) legally.

\subsection{Abuse}
Card holder making purchases on card that he/she has no intention of paying gives another type of fraud called abuse. Here card-bearer may pre-meditate+ activity just before they file for personal bankruptcy. These losses that are caused due to this ``bankruptcy fraud'' are a part of charge-off losses.
Causing adverse effects on business and society, and accounting for about billions of dollars of lost revenue each year \cite{5}, credit card frauds have become a major issue worldwide. While some statistics show about \$400 billion cost of a year, some other figures show about 1.6 billion pounds total yearly loss of UK insurers due to such fraudulent attempts. By 2005 with market splitting in two groups based on transaction types namely the credit group, that leads the market, including countries like Spain, Belgium, Italy, and Greece. With two countries, namely the United Kingdom and Ireland lacking competitors in terms of transaction product. Whereas, other group is of debit card users example Sweden. However, Germany being underserved by credit cards earlier, the market there was estimated to potentially increase by 23.3\% from 2004 to 2009, reaching an approximate of €56,477 million (Euromonitor International, 2006) \cite{38}.

The observed patterns in behavior of the customers regarding payments are related with the assumptions that customers use cards instead of cash (Euromonitor International, 2006). Now the loss is actually affecting everyone even if one hasn’t been defrauded, paying of credit and charge debts increases the cost of goods and services. 

Hence the need of the hour is a fraud Detection system that can distinguish an incoming transaction request as a fraudulent or non-fraudulent and hence alarm the banks \cite{32}. Various machine learning techniques can be used for prediction \cite{42}.

The rest of the paper is organized as: Section II consists of brief review of methodologies that have been proposed for a fraud detection system. The following Fig.1. depicts the methodologies that we have reviewed in this paper. Section III summarizes the techniques with their corresponding results and Section IV gives the conclusion.

\section{FRAUD DETECTION APPROACHES}
Machine learning finds its usage across the applications such as prediction, optimization, detection, classification etc.~\cite{x1,x2,x3,x4,x5,x6,x7,x8,x9,x10,x11,x12,x13,x14,39,40,41,42}.
\subsection{Hidden Markov Model(HMM)}
A Hidden Markov Model is a stochastic model that has a set of states that are finite with a set of transition possibilities and rate parameters for those transitions \cite{23}. The basis of the HMM Model is the Markov property that states that future events do not depend on the earlier states and only on the current ones. This property makes it useful in predictive modelling and probability forecasting. Detection of fraud using this model as specified in \cite{1}. In this, they model the human behavior based on card holders spending habits and a state in the model is the type of the purchase. Only three price ranges that is l-low, m-medium, h-high are considered that make 3 observation symbols. For instance, letting l= (0, \$200], m= [\$200, \$400], and h= [\$400, credit card limit]. Metrics used by this model are True Positive, TP- False Positive, FP and Accuracy. Accuracy that is proposed remains close to 80 percent for all inputs large or small. 

But when there is no information available about the profile, there is some degradation in performance in TP-FP metrics. And when there is slight difference between genuine and malicious transactions, then also the FDS suffers degradation in its performance by a decrease in number of TPs or a rise in FPs.

\begin{figure}
	\centering
	\includegraphics[width=\textwidth]{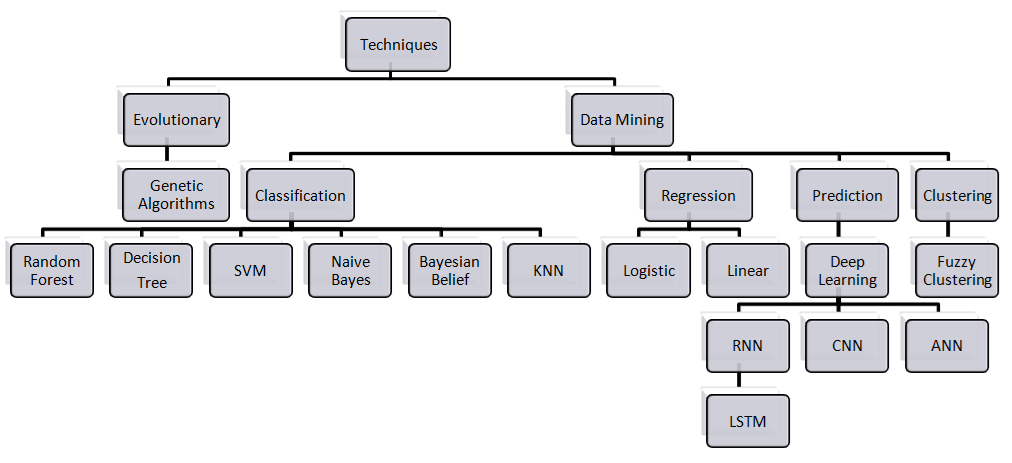}
	\caption{Taxonomy of fraud detection techniques}
	\label{fig:fig1}
\end{figure}

\subsection{Decision Tree}
A supervised learning algorithm \cite{14}, A decision tree which is in the form of tree structure, consisting of root node and other nodes split in a binary or multi-split manner further into child nodes with each tree using its own algorithm to perform the splitting process, until no more splitting is required that will make a difference in our model, associating each attribute with a value that is input variables related to the method that is being used as described by Y. Sahin and E. Duman \cite{3}. With the tree growing, there may be possibilities of overfitting of the training data with possible anomalies in branches, some errors or noise. Hence pruning is used for improving classification performance of the tree by removing certain nodes. Ease in the use, and the flexibility that the decision trees provide to handle different data types of attributes make them quite popular \cite{28}.

\subsection{Random Forests}
The instability in single trees and sensitivity to some training data led to development of another model that is random forests. With each tree being built independent of each other computational efficiency of random forest is comparatively better \cite{37}. It is basically an ensemble of regression and/or classification trees with it obtaining variance amongst its trees and hence are easy to use because of use of only two randomness sources or parameters that is building trees using trained data separate bootstrapped along samples with considering only a random data attribute subset to build each tree as specified \cite{5}.

\subsection{Bayesian Belief Networks}
This technique uses Bayes Theorem to compute the probability of a hypothesis and identify it to be true or false. A classifier is used to calculate the conditional probabilities for all the possible classes and insert it into the class that has the highest value of conditional probability for a particular value of X \cite{27}. Graphically, it is depicted in the form of a directed graph that is acyclic in nature, where the samples are represented by the nodes of the graph and the dependencies between them are reflected through the directed edges. Two variables are said to be independent if there are no connecting edges between them. It also gives a specification and factorization of the joint probability distribution \cite{8}.

\subsection{Genetic Algorithm}
This algorithm puts forth the idea of natural evolution, where the survival chances of the stronger or fitter individuals of the population is more than the weaker individuals of the population \cite{36}. The stronger members are chosen to reproduce and thus the mean fitness of the entire population improves. The fitter members of a certain generation are chosen as the parents for the next generation and the lesser fit members are discarded. It depends on many parameters like number of parents \& children, reproduction, fitness \& selection functions and some termination criteria \cite{11}.

\subsection{Logistic Regression}
It is an appropriate technique that can be used in predictive analysis when the dependent variable is dyadic or binary \cite{26}. Since the categorization of transactions being fraud is a double-edged variable, this technique can be used. This statistical classification model based on probabilities detects the fraud using logistic curve. Since the value of this logistic curve varies from 0 to 1, it can be used to interpret class membership probabilities. 

The dataset fed as input to the model is being classified for training and testing the model. Post model training, it is tested for some minimum threshold cut-off value for prediction. Then the most significant variables are selected and the model is tuned accordingly. The accuracy of prediction came out to be 70\%. Since the logistic regression, based on some threshold probabilities can divide the plane using a single line and divides dataset points into exactly two regions. Hence, the outlier points are not handled effectively \cite{14}. It uses natural logarithmic function to calculate probability and to show that the results fall under a particular category \cite{5}.

\subsection{Support Vector Machines}
Support vector machines or SVMs are linear classifiers as stated in \cite{5} that work in high dimensionality because in high-dimensions, a non-linear task in input becomes linear and hence this makes SVMs highly useful for detecting frauds. Due to its two most important features that is a kernel function to represent classification function in the dot product of input data point projection, and the fact that it tries finding a hyperplane to maximize separation between classes while minimizing overfitting of training data, it provides a very high generalization capability \cite{29}.

\subsection{K-Nearest Neighbours}
This is a supervised learning technique that achieves consistently high performance in comparison to other fraud detection techniques of supervised statistical pattern recognition \cite{25}. Three factors majorly affect its performance: distance to identify the least distant neighbors, some rule to deduce a categorization from k-nearest neighbor \& the count of neighbors to label the new sample. This algorithm classifies any transactions that occurred by computing the least distant point to this particular transaction and if this least distant neighbor is classified as fraudulent then the new transaction is also labeled as a fraudulent one. Euclidean distance is a good choice to calculate the distances in this scenario. This technique is fast and results in fault alerts. Its performance can be improved by distance metric optimization \cite{20}.

\subsection{Fuzzy Clustering}
Based on the past activities of the users or customers this technique helps determine normal usage patterns of the users \cite{35}. Whenever there is a deviation in the normal patterns, a suspicion score is calculated, accordingly the transactions are categorized as legitimate or fraudulent or suspicious. Further the suspicious transactions need to be recognized as fraudulent or occasional deflection from the usual patterns by the genuine customers itself. This is achieved by applying learning techniques through the neural networks, which effectively helps reduce false alarms \cite{19}. 

\subsection{Neural Networks}
A neural network is a network of neurons, comprising of perceptron which is a linear binary classifier which helps to classify the inputs data. Each perceptron consists of four parts: Input layer, Weights and Bias, Weighted Sum and Activation Function \cite{33}. Our analysis covers three different neural networks:

\subsubsection{Artificial Neural Network (ANN)} An ANN is composed of of three types of layers: input, hidden and output. The data to be trained travels from the input to intermediate hidden and then to the output layer, termed as Forward Propagation. The network is initially trained with normal behavior of the cardholder. The transactions that seems to be fraud are then Backpropagated through the network and are then classified as the fraudulent and non-fraudulent transactions. This technique was found to be quite efficient as a Neural Network doesn’t need to be reprogrammed so its processing speed is high \cite{24}.

\subsubsection{Convolution Neural Network (CNN)} CNN contains neurons with learnable values weights and biases with each neuron receiving various input vectors, and then calculate sum that is weighted over them, pass the value through an activation function and deliver an output~\cite{30,31}. Krishna and Reshma \cite{13} proposed a model in which they trained their model using CNN. Initially, feature selection algorithm was applied on the original data. Then, to overcome the imbalance in the data or the skewed distribution, SMOTE (Synthetic minoring oversampling technique) helps generating synthetic transactions that are fraud for balancing the dataset. The next step involved feature transformation and then the features were converted into a matrix which was given as input. They used soft-max activation function.

\subsubsection{Recurrent Neural Network (RNN)} RNNs are networks with loops in them having the ability to hold information. It can be thought of as a multilayer perceptron, where each of them passes an information to its successor \cite{34}. RNN has internal states which are updated after each unit time. Though, RNN being a powerful and simple model, still training them properly with gradient descent is hard. There are problems like Vanishing and Exploding Gradient Descent associated with it \cite{15}. These problems tell as the length of the input sequence grows, the gradient may either increase or decrease exponentially. These problems can be overcome using techniques like by removing the dependency of the sequence state vectors on the weight matrix and introducing memory cells which was termed as LSTM (Long Short-Term Memory) Network as proposed by Bayer \cite{17,40}.

\section{Summary}
Table 1 summarizes some of the significant solutions proposed for the fraud detection using machine learning. The study shows that the Fraud Bayesian Network Classifier followed by probability threshold is more beneficial than Naïve Bayes, Tree Augmented Naïve Bayes, Support Vector Machines and Decision Trees on the PagSeguro dataset when precision, recall and economic efficiency are taken into account \cite{16}. Bayesian Learning when taken with Dempster-Shafer Theory resulted in 98\% True Positives and less than 10\% False Positives \cite{7}. Genetic Algorithms along with Scatter Search improves the performance by 200\% when applied on the existing systems of a major bank of Turkey \cite{11}. Although ANN detects frauds faster, but Bayesian Belief is better as it is able to detect 8\% more frauds as recorded on data provided by Serge Waterschoot at Europay International \cite{8}. When dataset is highly imbalanced and independent of the rate of frauds Bagging Ensemble Classifier is one of the most stable approaches and has a high fraud catching rate \cite{9}. As observed on the German dataset using Big Data Analytical Framework with Hadoop it was observed that Random Forest Decision Tree outperforms Logistic Regression, Decision Trees and Decision Tree Random Forest in terms of precision and accuracy \cite{14}. Whereas Deep Networks training approach handled data granularity with high accuracy on the same dataset \cite{12}. Long Short-Term Memory improves upon transactions that are face to face, but is more likely to overfit i.e. layers have fewer nodes \cite{15}. The results on the Nation banks credit card warehouse showed that Decision Tree worked better than Support Vector Machines in terms of accuracy \cite{3}. The behaviour certificate model that was tested on the data generated using simulator performed well overall in comparison to Support Vector Machines \cite{10}. A lot of financial resources were saved when Cost-Sensitive Decision Tree approach was used. It also outperformes traditional classifiers in terms of total number of frauds detected \cite{6}. Convolutional Neural Networks along with SMOTE performs better than Neural Networks \cite{13}. The True Positives accounted for 93.9\% and False Positives around 6.10\% when Fuzzy Clustering and Neural Networks where implemented on data developed by Panigrahi \cite{19}. A real world credit card data by US bank put forth the better performance of Distributed deep learning than the non-privacy baseline approaches \cite{4}. Other important factors are further discussed in Table 1.

	\begin{tabularx}{\linewidth}{lXXlXX}
		\multicolumn{6}{c}{Table 1: A detailed comparison fraud detection approaches} \\
		\toprule
		Work & Technique used & Dataset used & Pre- Processing & Performance Metrics & Result \\
		\midrule
		\cite{16} & Bayesian Network Classifier (HHEA), instance reweighing and probability threshold analysis.&PagSeguro (Brazilian Online Payment Service)&$\checkmark$&HM between Precision and Recall, and Economic Efficiency.&Fraud BNC following Probability threshold - more beneficial than NB, TAN, SVM, Decision Trees, Logistic Regression	\\
		\cite{11}	&	Genetic Algorithm and Scatter Search.	&	Major Bank of Turkey (industrial partner).	&	$\checkmark$	&	Misclassification cost based on TP, TN, FP, FN, TFL, S, r., No. of frauds, Ratio of legitimate transactions, Class Imbalance	&	Improved existing performance by 200\%.	\\
		\cite{1}	&	Hidden Markov Model	&	NA	&	$\checkmark$	&	TP, FP	&	80\% accurate	\\
		\cite{8}	&	Bayesian Belief, ANN	&	Provided by Serge Waterschoot at Europay International	&	$\checkmark$	&	TP, FP	&	Bayesian Belief, better than ANN. 8\% more frauds detected. But ANN detects faster.	\\
		\cite{9}	&	Bagging Ensemble Classifier	&	Real world credit card dataset obtained from USCO-FICO competition.	&	$\checkmark$	&	Fraud Catching Rate, False Alarm Rate, Balanced Classification Rate, Mathews Correlation coefficient.	&	Stable, Fraud catching rate is high, Independent of rate of frauds, suitable for highly imbalanced dataset.	\\
		\cite{14}	&	Big Data Analytical Framework with Hadoop	&	German Dataset	&	$\checkmark$	&	FP, TP.	&	Random Forest Decision Tree performs best in terms of accuracy and precision among LR, DT \& DTRF.	\\
		\cite{15}	&	LSTM, State of art methods.	&	Dataset Recorded from March to May 2015.	&	$\checkmark$	&	Robustness against imbalance classes, Attention to business specific interests.	&	LSTM’s more accurate as compared to RF, improves upon face to face transactions, but LSTM is prone to overfit (layers have fewer nodes).	\\
		\cite{12}	&	Deep Networks by training a deep network	&	German Credit Data	&		&	Accuracy, Variance	&	High Accuracy handing data granularity	\\
		\cite{5}	&	Logistic Regression, SVM, Random Forests	&	From international credit card operation of study in ANN tuned by Genetic Algorithm (GAs) to detect fraud	&		&	Accuracy, sensitivity, Specificity, precision, F-measure, G-mean, wtdAcc.	&	Random Forest- better performance overall. Logistic Regression- better performance through different datasets.	\\
		\cite{3}	&	Decision Trees, SVM	&	Nation banks credit card warehouse	&	$\checkmark$	&	Accuracy	&	Decision tree outperforms SVM	\\
		\cite{4}	&	Distributed deep learning	&	real-world credit card data by a US bank	&	$\checkmark$	&	AUC comparison.	&	Better performance than non-privacy baseline.	\\
		\cite{10}	&	Behavior certificate model	&	Generated using simulator	&	$\checkmark$	&	Accuracy, Recall, Specificity, precision, F-measure, G-mean	&	Overall better performance than SVM	\\
		\cite{7}	&	Dempster-Shafer theory and Bayesian learning	&	Transaction history repository (THD) built using simulator	&		&	FP-TP	&	Upto 98\% TP and less than 10\% FP	\\
		\cite{6}	&	Cost-sensitive Decision Tree	&	Banks credit card data warehouses	&	$\checkmark$	&	Saved Loss Rate (SLR)	&	Saved much more financial resources, outperforms traditional classifier in number of frauds detected.	\\
		\cite{13}	&	Convolutional Neural Network	&	Credit card fraud data	&	$\checkmark$	&	TP-FP, FN-TN, Precision and Recall with their HM. 	&	With SMOTE, outperforms NN.	\\
		\cite{19}	&	Fuzzy clustering and neural networks	&	Developed by Panigrahi \cite{7}	&		&	FP, TP/Sensitivity, TN/Specificity	&	Upto 93.9\% TP and less than 6.10\% FP	\\
		\cite{21}	&	K-Nearest Neighbor	&	Real data from private bank	&	$\checkmark$	&	Recall, F-measure, Specificity, Accuracy, Precision, 	&	Performance is evaluated on the basis of the metrices.	\\
		\bottomrule
	\end{tabularx}

\section{Conclusions}
We have come across various fraud detection techniques that exists today but none of them were competent enough to detect the fraud at the time it actually took place. They detected the frauds which happened in the past. The setback of all the techniques discussed so far give accurate results only when performed on a particular dataset and sometimes with some special features only. But we need to establish a technology that works equally precisely and accurately under all circumstances and with various datasets. Techniques like SVM works better than Logistic Regression when there is a class imbalance and comparatively Random Forest performs better among all three \cite{5,14}.

Bagging Ensemble Classifier is suitable for highly imbalanced dataset \cite{9}. Some techniques like Decision Tress and SVM gives better results on raw unsampled data whereas techniques like ANN and Bayesian Belief Network have high accuracy and detection rate but are expensive to train \cite{8}.  Similarly, KNN and SVM gives better results with small datasets but are not preferable with large datasets.

Among all the techniques discussed, it is found that Neural Networks detects frauds with high precision and performs best. But as they are expensive to train and can also be over trained in case of fewer nodes as in case of LSTM \cite{15}. So, in order to minimize this cost, we can pair neural network with some augmentation techniques like Genetic Algorithms or Artificial Immune Systems by selecting optimized weight edges and eliminating those weights that causes error.



\end{document}